%% file: sample-sigconf.tex
\documentclass[sigconf]{acmart}
\pdfoutput=1

\usepackage{booktabs} 
\usepackage{color}
\usepackage{threeparttable}
\usepackage{diagbox}

\usepackage[ruled]{algorithm2e} 

\setcopyright{rightsretained}

\acmDOI{10.475/123_4}

\acmISBN{123-4567-24-567/08/06}

\acmArticle{4}
\acmPrice{15.00}

\editor{Jennifer B. Sartor}
\editor{Theo D'Hondt}
\editor{Wolfgang De Meuter}

\begin{document}
\title[Solving Pictorial Jigsaw Puzzle by Collective Intelligence]{Solving Pictorial Jigsaw Puzzle by Stigmergy-inspired Internet-based Human Collective Intelligence}

\author{Bo Shen}
\orcid{1234-5678-9012}
\affiliation{%
  \institution{Peking University, Beijing, China}
  \streetaddress{P.O. Box 1212}
  \city{Beijing}
  \country{China}
  \postcode{100871}
}
\email{shenbo@pku.edu.cn}
\author{Wei Zhang}
\orcid{1234-5678-9012}
\affiliation{%
  \institution{Peking University, Beijing, China}
  \streetaddress{P.O. Box 1212}
  \city{Beijing}
  \country{China}
  \postcode{100871}
}
\email{zhangw.sei@pku.edu.cn}

\author{Haiyan Zhao}
\orcid{1234-5678-9012}
\affiliation{%
  \institution{Peking University, Beijing, China}
  \streetaddress{P.O. Box 1212}
  \city{Beijing}
  \country{China}
  \postcode{100871}
}
\email{zhhy.sei@pku.edu.cn}

\author{Zhi Jin}
\orcid{1234-5678-9012}
\affiliation{%
  \institution{Peking University, Beijing, China}
  \streetaddress{P.O. Box 1212}
  \city{Beijing}
  \country{China}
  \postcode{100871}
}
\email{zhijin@pku.edu.cn}

\author{Yanhong Wu}
\orcid{1234-5678-9012}
\affiliation{%
  \institution{Peking University, Beijing, China}
  \streetaddress{P.O. Box 1212}
  \city{Beijing}
  \country{China}
  \postcode{100871}
}
\email{wyh@pku.edu.cn}

\renewcommand{\shortauthors}{BoShen et al.}

\begin{abstract}
The pictorial jigsaw (PJ) puzzle is a well-known leisure game for humans. Usually, a PJ puzzle game is played by one or several human players face-to-face in the physical space. In this paper, we focus on how to solve PJ puzzles in the cyberspace by a group of physically distributed human players. We propose an approach to solving PJ puzzle by stigmergy-inspired Internet-based human collective intelligence. The core of the approach is a continuously executing loop, named the \emph{EIF} loop, which consists of three activities: exploration, integration, and feedback. In \emph{exploration}, each player tries to solve the PJ puzzle \emph{alone}, without direct interactions with other players. At any time, the result of a player's exploration is a partial solution to the PJ puzzle, and a set of rejected neighboring relation between pieces. The results of all players' exploration are integrated in real time through \emph{integration}, with the output of a continuously updated \emph{collective opinion graph} (COG). And through \emph{feedback}, each player is provided with personalized feedback information based on the current COG and the player's exploration result, in order to accelerate his/her puzzle-solving process. Exploratory experiments show that: (1) supported by this approach, the time to solve PJ puzzle is nearly linear to the reciprocal of the number of players, and shows better scalability to puzzle size than that of face-to-face collaboration for 10-player groups; (2) for groups with 2 to 10 players, the puzzle-solving time decreases 31.36\%-64.57\% on average, compared with the best single players in the experiments.
\end{abstract}

%
%

\ccsdesc[500]{Human-centered computing~Computer supported cooperative work}
\ccsdesc[500]{Human-centered computing~Web-based interaction}
\ccsdesc[100]{Human-centered computing~Asynchronous editors}

\keywords{Collective intelligence, stigmergy, human-computer interaction, pictorial jigsaw puzzle, complex problem solving}

\maketitle

\input{samplebody-conf}


\end{document}

%% file: samplebody-conf.tex
\pdfoutput=1
\section{Introduction}

Pictorial jigsaw (PJ) puzzle is a well-known leisure game for people from children to adults. In a PJ puzzle game, the goal is to recover an image with human-sensitive contents from $n$ different pieces of the image, as fast as possible. Besides the appearance of a leisure game, PJ puzzle has a deep metaphorical meaning. It embodies perfectly a kind of complex problems that can not be resolved in a top-down manner, but only in a bottom-up, exploring and growing manner. That is, such a problem usually can not be pre-decomposed into a set of simpler sub-problems and then be resolved by synthesizing solutions to the sub-problems. The problem solver has to explore a large (even open) set of possible information pieces, find different ways to combine these pieces, and make one or several partial solutions continually grow by integrating more and more pieces until finding an acceptable solution. This kind of complex problems can be found in many research and practical fields, including geology \cite{torsvik2003rodinia}, medicine \cite{burkitt1975large}, biology \cite{marande2007mitochondrial} \cite{brimacombe1995structure}, sociology \cite{verweij2006clumsy}, and complex information synthesis \cite{Hahn2016} \cite{Chang2016}. 

Usually, a PJ puzzle game is played by one single human player, or by several human players face-to-face in the physical space. Motivated by the metaphorical meaning of PJ puzzle, in this paper, we focus on the problem of how to efficiently solve PJ puzzle in the cyberspace by a group of physically distributed human players. In particular, we propose an approach to solving PJ puzzle by stigmergy-inspired Internet-based human collective intelligence. 

When it is firstly proposed, \emph{stigmergy} \cite{theraulaz1999brief} \cite{karsai1999decentralized} \cite{susi2001social} \cite{dorigo2000ant} denotes a kind of mechanism about environment-mediated indirect interaction between insects, and makes it possible to explain the seemingly paradoxical phenomena observed in an insect colony: individual insects work as if they were alone, whereas their collective behaviors appear to be well coordinated. These phenomena are usually called \emph{collective intelligence} (CI) or \emph{swarm intelligence} (SI) in social insects. From the viewpoint of \emph{stigmergy}, the CI phenomenon of an insect colony emerges in the following way: individuals leave their traces in or made modifications to the environment; then traces and modifications are perceived by individuals (others or themselves) in the colony, and trigger them to further leave new traces in or make new modifications to the environment; therefore, individual behaviors coordinate with each other and form a positive feedback loop, leading to the appearance of intelligent self-organizing collective behaviors.

The concept of stigmergy reveals a key component in CI, namely, the \emph{environment}. It is the environment (in particular, the dynamics of the environment) that enables the environment-mediated large-scale indirect interactions among individuals in a group. One important characteristic of stigmergy-enabled CI is the good scalability of collaboration: a large scale group of individuals can participant in the environment-mediated collaboration, without sacrificing the group's and any individual's working efficiency. 

Guided by the concept of stigmergy, one of the keys to design an artificial stigmergy-enabled CI system is to construct an artificial environment that matches the characteristics of the problem to be resolved. We think that the environment in stigmergy should undertake two kinds of responsibility. The first is \emph{integration}: the environment should integrate all the information pieces provided by individuals in the group into a well-structured collective-level artifact. The second is \emph{feedback}: the environment should provide personalized feedback information for each individual in the group based on the collective-level artifact and the individual's characteristics. In those natural stigmergy-enabled CI phenomena occurring in the physical space, the integration responsibility is carried out by physical laws. For example, it is the physical properties of pheromone that enable pheromone placed by different insects at the same location to be merged together, increasing the density of pheromone at that location. And in these natural CI phenomena, the feedback information for each individual is personalized according to the individual's location in the physical environment. But in the cyberspace, physical laws do not have direct effects on information processing, and physical locations also become meaningless. To develop an artificial stigmergy-enabled CI system in the cyberspace, we need to construct a problem-specific virtual environment with integration and feedback mechanisms suitable for the cyberspace.

\begin{figure}[!htp]
\includegraphics[width=1\linewidth]{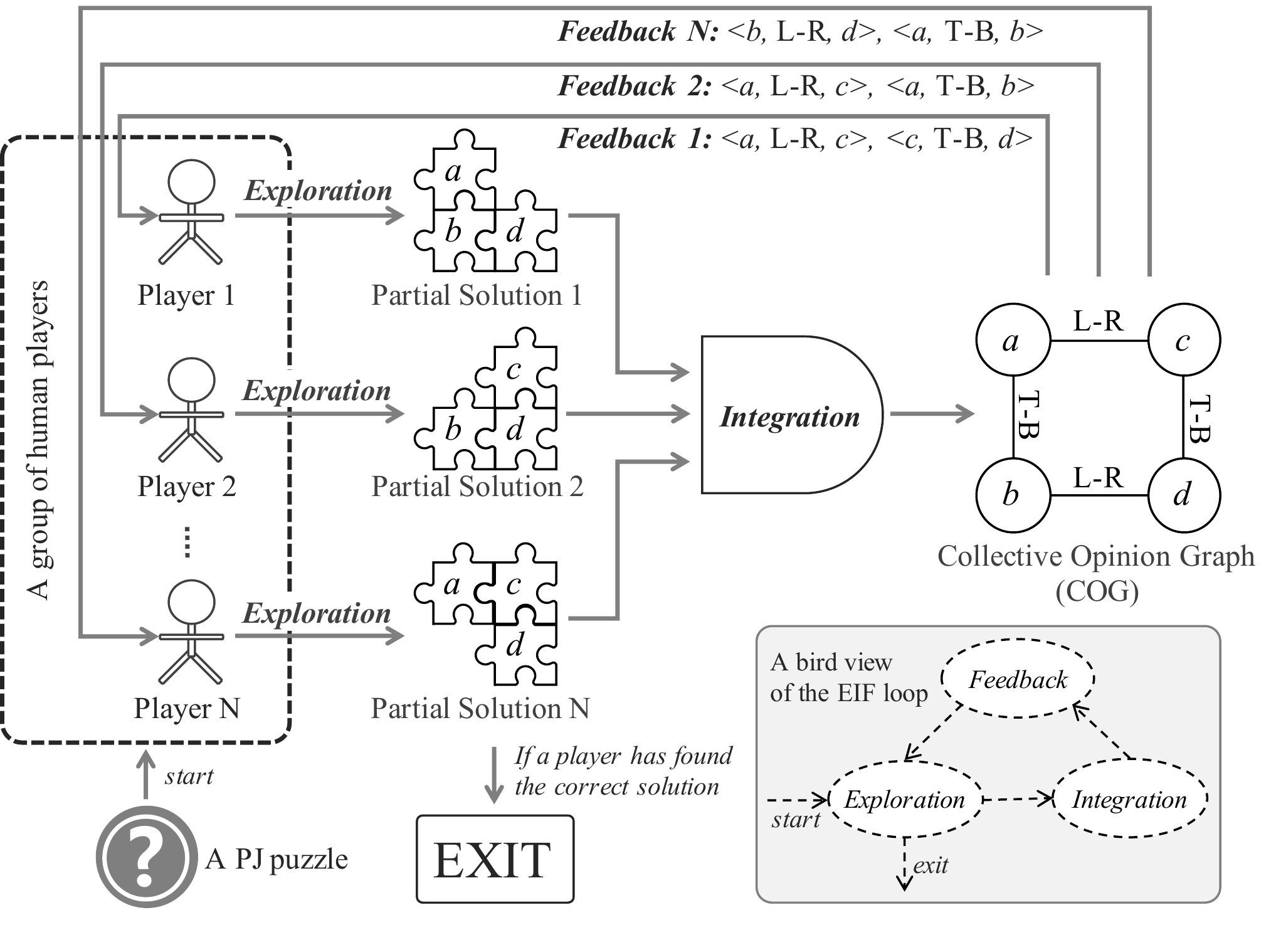}
\caption{An illustration of using the EIF loop to solve PJ puzzle by a group of physically distributed human players.}
\label{fig:eif}
\end{figure}

The core of the proposed approach in this paper is a continuously executing loop, named the \emph{EIF} loop, which consists of three asynchronously connected activities: exploration, integration, and feedback (see Fig.~\ref{fig:eif}). In \emph{exploration}, each player tries to solve the PJ puzzle alone (\emph{not really alone}), without direct interactions with other players. At any time, the result of a player's exploration is a partial solution to the PJ puzzle, as well as a set of rejected neighboring relation between image pieces. The results of all players' exploration are integrated in real time through \emph{intergation}, with the output of a continuously updated \emph{collective opinion graph} (COG). And through \emph{feedback}, each player continuously receives personalized feedback information based on the current COG and the player's current result, in order to accelerate her/his puzzle-solving process. Each player independently decides whether to accept or reject the feedback information. When any player find the correct solution, it means that the PJ puzzle is resolved and then the EIF loop will be terminated. 

The EIF loop is a refinement of stigmergy for human CI in the cyberspace. In particular, the EIF loop refines stigmergy in two points. First, the integration in EIF is not carried out by physical laws as in physical-space based stigmergy, but by human-defined software and algorithm, which gives sufficient flexibility when applying stigmergy to different artificial problem-solving situations. Second, instead of basing on an individual's physical location, the feedback in EIF depends on the current collective-level artifact and an individual's current exploration result. In addition, the feedback in physical-space based stigmergy is a kind of passive feedback, while in EIF, the feedback is active; that is, related information is automatically pushed to an individual.

We have implemented an on-line platform, called \emph{Crowd Jigsaw Puzzle} \footnote{Available at http://www.pintu.fun, and currently only supports the Chrome browser.}, to demonstrate the proposed approach. Based on this platform, we have conducted a set of controlled experiments to investigate the feasibility and effectiveness of the proposed approach. Our experiments show that:
\begin{enumerate}
\item Given a PJ puzzle, supported by this approach, the time to solve PJ puzzle is nearly linear to the reciprocal of the number of players, and shows a better scalability to puzzle size than that of resolving PJ puzzle by face-to-face collaboration. 
\item In particular, for groups with 2 to 10 players, the puzzle-solving time decreases 31.36\%-64.57\% on average, compared with the best single players involved in the experiments.
\end{enumerate}

The main contributions of this paper include four points.
\begin{enumerate}
\item A general model inspired by stigmergy to enable human collective intelligence in solving PJ puzzle. We name this model the EIF (\emph{exploration-integration-feedback}) loop. 
\item A structured way to integrate each player's opinion in the player group when solving PJ puzzle. We name the output of integration the \emph{collective opinion graph} (COG).
\item An on-line platform that supports a group of human players to solve PJ puzzle in a collaborative and decentralized way.
\item An empirical quantitative model that shows the cause-and-effect relation from the two factors of group size and puzzle size to the collective performance of solving PJ puzzle.
\end{enumerate}

The rest of the paper is structured as follows. Section 2 introduces the related work of the proposed approach. Section 3 presents the stigmergy-inspired approach to solving PJ puzzle by a group of physically-distributed players. Section 4 evaluates the feasibility and effectiveness of the proposed approach through a set of controlled experiments. Section 5 discusses some elementary problems related to the proposed approach, and highlights our future work. Finally, Section 6 concludes this paper with a short summary.

\section{Related Work}

\subsection{Automatic Solvers for Jigsaw Puzzle}

As a formal research problem, jigsaw puzzle was introduced into the academic community in 1964 by Freeman and Garder \cite{freeman1964apictorial}. In their research, the two scholars focused on apictorial jigsaw puzzle (i.e., puzzle in which all pieces contain no chromatic but shape information), and proposed a pattern-recognition based automatic approach to solving a special kind of apictorial jigsaw puzzle. 

The essential difficulty of jigsaw puzzle has been formally investigated. Berger \cite{berger1966undecidability} proved that a general edge-matching puzzle can not be solved by any algorithms in general. Demaine et al. \cite{demaine2007jigsaw} proved that three kinds of jigsaw puzzle (i.e., classic apictorial jigsaw puzzle, edge-matching puzzle, and polyform packing puzzle) are all NP-complete, and can be converted with each others.

In general, research on automatic solvers for jigsaw puzzle can be largely classified into three categories. The first category focuses on apictorial jigsaw puzzle \cite{Goldberg2002} \cite{Kong2001} \cite{Radack1982} \cite{Wolfson1988}. In this category, apictorial jigsaw puzzles are treated as computational geometry problems \cite{Goldberg2002}. The second category adds chromatic information into pieces in apictorial jigsaw puzzle \cite{Kosiba1994} \cite{Makridis2006} \cite{Nielsen2008} \cite{Yao2003}, which decreases the difficulty of jigsaw puzzle in some degree since additional clues are provided. The third category focuses on pictorial jigsaw (PJ) puzzle \cite{cho2010probabilistic} \cite{pomeranz2011fully} \cite{Yang2011} \cite{gallagher2012jigsaw} \cite{sholomon2013genetic} \cite{sholomon2016automatic} \cite{Son2014}. In a PJ puzzle, all pieces have identical shapes, and if correctly composed, all pieces together will show an image with human-sensitive contents. PJ puzzle is usually treated as computer vision problems \cite{sholomon2013genetic} \cite{sholomon2016automatic}.

Most of the existing automatic solvers for PJ puzzle include two basic components: (1) a \emph{compatibility metric} to quantitatively evaluate how well two pieces fit together, and (2) an \emph{assembly strategy} to find high-quality candidate solutions. 
\begin{itemize}
\item For example, in the state-of-the-art solver for PJ puzzle with known piece-orientation \cite{sholomon2013genetic} \cite{sholomon2016automatic}, the compatibility metric is calculated by summing the squared color difference along the abutting edges of two pieces; this metric is proposed by Cho et al. \cite{cho2010probabilistic}. And the assembly strategy is incarnated in a genetic algorithm, in which, a crossover operator is defined to obtain a child solution from two parents, partly using the \emph{best-buddy} metric proposed by Pomeranz et al. \cite{pomeranz2011fully}. 
\item For another example, in the state-of-the-art solver for PJ puzzle with unknown piece-orientation \cite{Son2014}, the compatibility metric called \emph {MGC} is used, which is proposed by Gallagher \cite{gallagher2012jigsaw}. And the assembly strategy is incarnated in a two-step algorithm: step (1) bottom-up recovering puzzle loops with different dimensions in the puzzle, and step (2) top-down merging puzzle loops identified in the first step.    
\end{itemize}

It should be pointed out that, the effectiveness of the two state-of-the-art PJ solvers heavily depend on the following two assumptions:

\begin{itemize}
\item There is a high positive correlation between a high compatibility value of two pieces and the fact that the two pieces are adjacent in the correct solution. This assumption is wrong in general, since we can easily construct a PJ puzzle that breaks this assumption (i.e., the edge-matching puzzle \cite{demaine2007jigsaw}).
\item The compatibility value of two pieces can be fully derived from the outermost one or two columns/rows of pixels in pieces. This assumption is counterintuitive from the viewpoint of human players, since they depend much on the contents of two pieces to evaluate the compatibility value. In our experiments, we erased the outermost 2 columns/rows of pixels of each piece, but no players even notice this fact.
\end{itemize}

As far as we know, our work is the first attempt to solve PJ puzzle by stigmergy-inspired Internet-based human collective intelligence. However, our goal is not to show that our approach can outperform automatic solvers for any kinds of PJ puzzle; for those PJ puzzles satisfying the two assumptions above, the state-of-the-art solvers have done well. What we are really interested in is the kind of complex problems that are embodied by jigsaw puzzle in general. Our long-term goal is to show that human beings, if being organized appropriately, do possess some distinctive capabilities that can not be easily substituted by machines. 

\subsection{Stigmergy and Collective Intelligence}

Many scholars use the concept of stigmergy to explain human CI phenomena or design artificial human CI systems. Parunak \cite{parunak2005survey} pointed out that stigmergy is not a mechanism specific to insects, and human-human stigmergy is pervasive in both pre-computer and computer-enabled social systems. Furthermore, Parunak \cite{parunak2005survey} proposed a general architecture of stigmergy, consisting of four basic components: agent's state, agent's dynamics, environment's state, and environment's dynamics. Heylighen \cite{heylighen1999collective} showed how to design algorithms to develop a collective mental map of an Internet-based human group to resolve complex problems that can not be well resolved by individuals. Here, the algorithms correspond to the environment's dynamics, and the collective mental map to the environment's state, in the architecture of stigmergy proposed by Parunak \cite{parunak2005survey}. Malone et al. \cite{malone2010collective} reported a set of reusable elements in those successful Internet-based CI phenomena, elements about how information pieces provided by individuals are integrated, and elements about how to stimulate individual's involvement. 

We have observed two successful artificial CI systems in the cyberspace that perfectly embodies the EIF loop proposed in this paper. One is the \emph{UNU} system \cite{Rosenberg2016} \cite{Rosenberg15}, an online platform that enables a large group of players to solve a question by collectively selecting an answer from a set of candidates. In the virtual environment of UNU, the question-solving process can be understood as an instance of the EIF loop: (1) each player shows her/his opinion by placing a virtual magnet at an appropriate position relative to a virtual puck (i.e., \emph{exploration}); (2) all the forces exerted on the puck from magnets are integrated into a single force, making the puck move in a specific direction (i.e., \emph{integration}); (3) the puck's movement is observed by all the players (i.e., \emph{feedback}), and stimulates them to adjust the positions of their magnets accordingly (i.e., \emph{exploration} again). The second is the \emph{EteRNA} system \cite{Lee2014}, a multi-player on-line game that engages non-scientists in solving protein structure problems. The problem-solving process in EteRNA can also be viewed as an instance of the EIF loop: (1) players design their solutions in their own workspaces (i.e., \emph{exploration}); (2) players review and vote for the best solutions, and a set of top-voted solutions are synthesized and verified by chemical measurements (i.e., \emph{integration}); (3) the results are published on-line (i.e., \emph{feedback}), and stimulate players to start the next cycle of problem solving. 

Many scholars also investigate CI without using stigmergy. Levy and Bononno \cite{levy1997collective} discussed the ideal form of CI in the cyberspace and analyzed its influence on human societies. Woolley et al. \cite{woolley2010evidence} showed the evidence of CI in small face-to-face human groups in the physical space and proposed a quantitative factor to evaluate CI. Nielsen \cite{Nielsen2011} thought that the nature of CI in the cyberspace is a kind of \emph{designed serendipity}: from a large group of individuals, it is more possible to find a group of people who collectively possess the right information to resolve a problem. Maleszka and Nguyen \cite{Maleszka2015} focused on one characteristic of CI (that is, the whole is greater than the sum of its parts), and propose a mathematical model to integrate knowledge with hierarchical structures in order to find new knowledge that doesn't exist in the knowledge being integrated. 

\subsection{Crowdsourcing}

As a model for problem solving \cite{brabham2008crowdsourcing}, the term \emph{crowdsourcing} can be understood in two senses. In the \emph{narrow} sense, crowdsourcing represents ``\emph{the act of a company or institution taking a function once performed by employees and outsourcing it to an undefined (and generally large) network of people in the form of an open call}'' (or more concisely, the act of ``\emph{outsourcing to the crowd}'')\cite{howe2006crowdsourcing} \cite{howe2006rise}, which is the original meaning the term is coined to represent. In the \emph{broad} sense, crowdsourcing covers any activities that utilize the crowd's capability to resolve problems \cite{doan2011crowdsourcing}, which greatly generalizes the term's original meaning, making it a near-synonym for CI \cite{estelles2012towards}. 

Currently, there are two kinds of dominant crowdsourcing practice. The first kind focuses on resolving the problem that manifests itself as or can be easily decomposed into a large set of easy-for-human but difficult-for-computer tasks (usually called \emph{micro}-tasks) \cite{kittur2008crowdsourcing} \cite{ValentineRTRDB17}. The second kind focuses on resolving puzzles that are indecomposable or not easy to decompose: an institution makes an open call for the puzzle in a public media to attract experts, and offers an award to whomever comes up with the best solution. 

Compared with the two kinds of prevailing crowdsourcing practice, our approach can be viewed as a kind of more intelligent crowdsourcing, at the following two points. First, in our approach, there is no need to pre-decompose the to-be-resolved problem into a set of sub-tasks, since the problem will be resolved in a bottom-up manner. Second, our approach depends much on the self-organized collaboration among a group of individuals, while in the two kinds of crowdsourcing practice, collaboration is relatively weak. Kittur et al. \cite{Kittur2013} have pointed out that collaboration will be one of the elementary characteristics in the future of crowd work. Furthermore, if we understand crowdsourcing in the \emph{broad} sense, then artifical CI systems like UNU and EteRNA can also be viewed as crowdsourcing systems, in which collaboration becomes a key factor. In this sense, the two kinds of crowdsourcing practice are just a rudimentary form of artificial CI systems.

\section{Methodology}

In this section, we first introduce a special kind of PJ puzzle focused in our approach, and then describe the key concepts related to the \emph{exploration}, \emph{integration}, and \emph{feedback} activities when using the EIF loop to resolve PJ puzzle. 

\subsection{The Focused PJ Puzzle}
\label{sec:pjpuzzle}

There are many variants of jigsaw puzzle \cite{freeman1964apictorial} \cite{gallagher2012jigsaw}, according to the shape and chromatic information of pieces, whether the puzzle size or the orientation of each piece is known, whether the complete picture is known, and other related information. 

In this paper, we focus on a special kind of PJ puzzle that has the following characteristics.
\begin{enumerate}
\item The complete picture is a rectangle image with human-sensitive contents.
\item The complete picture is unknown for the players.
\item The puzzle size $M \times N$ (i.e. the number of pieces in row and column) is unknown for the players.
\item There is no overlap between any two pieces.
\item All pieces in a PJ puzzle have jagged-square (with rounded tab or slot on the four sides) shapes with the same size.
\item The orientation of each piece is fixed and same with its orientation in the complete picture.
\end{enumerate}

\begin{figure}[!htp]
\includegraphics[width=1\linewidth]{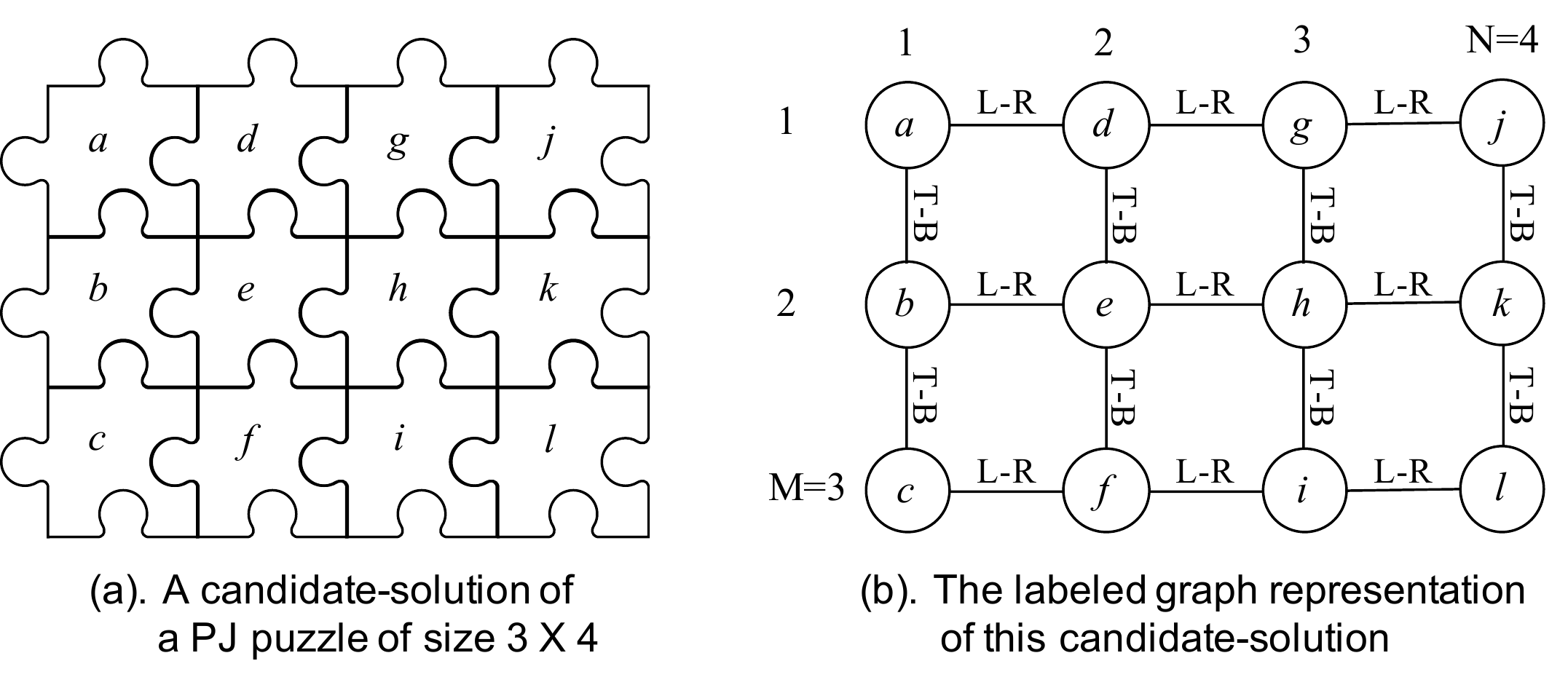}
\caption{A candidate solution to a PJ puzzle and the corresponding labeled graph representation.}
\label{fig:cs}
\end{figure}

In our approach, we use labeled graphs to represent candidate solutions or players' partial solutions (see Section~\ref{sec:exploration}) of PJ puzzle. Fig.~\ref{fig:cs} (a) shows a candidate solution to a PJ puzzle of size 3$\times $4, and Fig.~\ref{fig:cs} (b) shows the corresponding labeled graph representation. In the labeled graph, each vertex represents a distinct piece in the PJ puzzle, and each edge represents a neighboring relation between two pieces. There are two kinds of edge label: the L-R (\emph{left-right}) label, and the T-B (\emph{top-bottom}) label. A L-R labeled edge $\text{<}a,\text{L-R},d\text{>}$ means the right-side neighbor of $a$ is $d$ (or, the left-side neighbor of $d$ is $a$). A T-B labeled edge $\text{<}a,\text{T-B},b\text{>}$ means the bottom-side neighbor of $a$ is $b$ (or, the top-side neighbor of $b$ is $a$).

It should be pointed out that the human players of PJ puzzle are agnostic to the labeled graph representation: it is only used by the virtual environment of collective PJ puzzle solving, for information integration and feedback. 

\begin{definition}[Candidate Solution to a PJ Puzzle] \label{def:cs} Given a PJ puzzle of size $M \times N$, a candidate solution to this puzzle is a graph $S = (\mathcal{V}(S), \mathcal{E}(S))$ that satisfies the following two conditions.

\begin{enumerate}
\item $|\mathcal{V}(S)| = M \cdot N$.
\item There exists a row-column based encoding to all vertices in $\mathcal{V}(S)$, that is $\mathcal{V}(S) = \{ v_{ij} | 1 \le i \le M, 1 \le j \le N\}$, satisfying $\mathcal{E}(S) = \{ \text{<}v_{ij}, \text{L-R}, v_{i(j+1)}\text{>} \ | \ 1 \le i \le M, 1 \le j < N\} \cup \{\text{<}v_{ij}, \text{T-B}, v_{(i+1)j}\text{>} \ | \ 1 \le i < M, 1 \le j \le N \}$.
\end{enumerate}
\end{definition}

Given a PJ puzzle, the following notations are used in subsequent sections.

\begin{itemize}
\item $\mathcal{P}$: the group of human players.
\item $\mathcal{V}$: the vertex set consisting of all pieces in the puzzle.
\item $\mathcal{S}$: the set of all candidate solutions of the puzzle.
\item $\mathcal{E}$: the set of all edges appearing in any element in $\mathcal{S}$.
\item $tag(e)$: the label of an edge $e \in \mathcal{E}$.
\item $e.L, e.R$: the vertex playing the L and R role, respectively, in a L-R edge $e \in \mathcal{E}$.
\item $e.T, e.B$: the vertex playing the T and B role, respectively, in a T-B edge $e \in \mathcal{E}$.
\end{itemize}

\begin{proposition} Given a PJ puzzle of size $M \times N$, the following propositions are true: (1) $|\mathcal{S}| = (MN)!$; (2) $\forall S \in \mathcal{S} \cdot |\mathcal{E}(S)| = 2MN\text{-}M\text{-}N$;  and (3) $|\mathcal{E}| = 2MN(MN\text{-}1)$. 
\label{pro:cs}
\end{proposition}


That is, given a PJ puzzle of size $M \times N$, (1) there are totally $(MN)!$ candidate solutions, (2) each candidate solution contains $2MN\text{-}M\text{-}N$ neighboring relations between pieces, and (3) there are totally $2MN(MN\text{-}1)$ neighboring relations in all the candidate solutions. Solving the puzzle means locating the correct solution from $(MN)!$ candidate solutions, or locating the $2MN\text{-}M\text{-}N$ correct neighboring relations from $2MN(MN\text{-}1)$ candidate neighboring relations.

\subsection{Exploration}
\label{sec:exploration}

In the exploration activity, each human player tries to resolve the PJ puzzle \emph{alone} \footnote{Players are not really alone. There are indirect interactions between players, supported by the integration and feedback activities (see the next two subsections).} in her/his own workspace, without any direct interaction with other players. At any time, the result of a player's exploration consists of two artifacts: a partial solution to the puzzle, and a rejected edge set (that is, a set of rejected neighboring relations between pieces). 

\begin{definition}[A Player's Partial Solution at Any Time] \label{def:working-result} Given a PJ puzzle, for any player $p \in \mathcal{P}$, at time $t$, $p$'s partial solution to the puzzle is a set of connected graphs, denoted as $\mathcal{PS}(p,t)$, satisfying the following six conditions.

\begin{enumerate}
\item $\forall C \in \mathcal{PS}(p,t) \cdot \mathcal{V}(C) \subseteq \mathcal{V} \land \mathcal{E}(C) \subset \mathcal{E}$.
\item $\forall v \in \mathcal{V} \cdot \exists C \in \mathcal{PS}(p,t) \cdot v \in \mathcal{V}(C)$.
\item $\forall C_0, C_1 \in \mathcal{PS}(p,t) \cdot C_0 \neq C_1 \Rightarrow \mathcal{V}(C_0) \cap \mathcal{V}(C_1) = \varnothing$.
\item $\forall C \in \mathcal{PS}(p,t) \cdot \forall e, f \in \mathcal{E}(C) \cdot ((e \ne f) \land (tag(e) = tag(f))) \Rightarrow (((tag(e) = \text{L-R}) \land (e.L \ne f.L) \land (e.R \ne f.R)) \lor ((tag(e) = \text{T-B}) \land (e.T \ne f.T) \land (e.B \ne f.B)))$.
\item $\forall C \in \mathcal{PS}(p,t) \ \cdot$ there exists no loop consisting of edges with the same label.
\item $\forall C \in \mathcal{PS}(p,t) \ \cdot$ no other edges can be deduced from edges in $\mathcal{E}(C)$. 
\end{enumerate}
\label{def:ps}
\end{definition}

Condition 1 says that a connected graph should be formed by vertices in $\mathcal{V}$ and edges in $\mathcal{E}$. Condition 2 and 3 say that all the connected graphs in $\mathcal{PS}(p,t)$ form a \emph{partition} to vertices in $\mathcal{V}$. Condition 4 says that for any vertex in a player's partial solution, at each of the four sides (i.e., \emph{left}, \emph{right}, \emph{top}, \emph{bottom}) of the vertex, at most one edge can be connected. Condition 5 says that, for example, given three vertices $i,j,k$, it is impossible for a connected graph in $\mathcal{PS}(p,t)$ to include the three edges of $\text{<}i,\text{L-R},j\text{>}$, $\text{<}j,\text{L-R},k\text{>}$, and $\text{<}k,\text{L-R},i\text{>}$ simultaneously. Condition 6 says that if an edge can be deduced from the existing edges of a connected graph, then the deduced edge should also be added to this connected graph. 

\begin{figure}[!htp]
\includegraphics[width=0.9\linewidth]{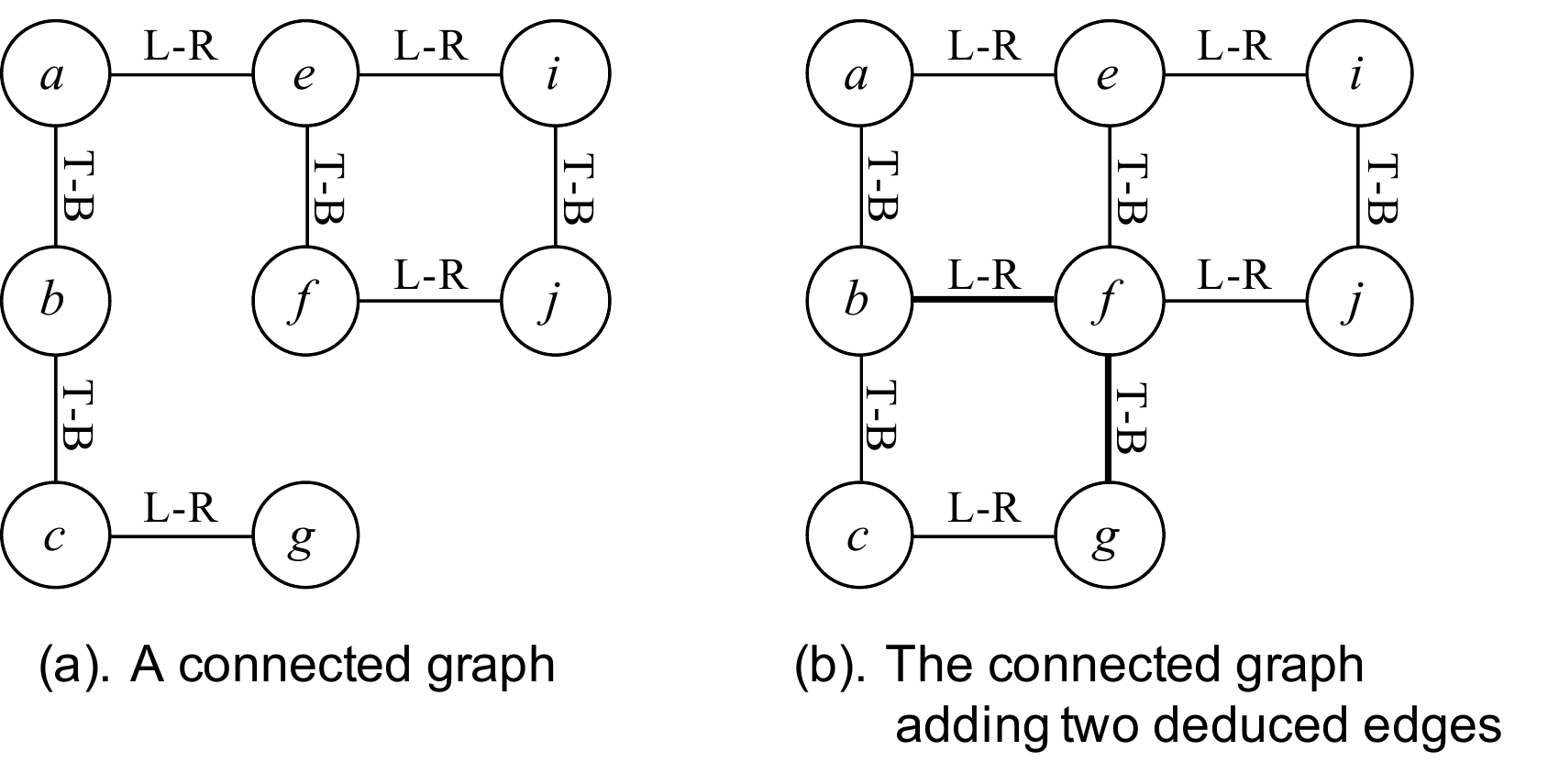}
\caption{Deducing edges from existing edges.}
\label{fig:deduced-edges}
\end{figure}

Fig.~\ref{fig:deduced-edges} shows an example of how edges are deduced from existing edges. From edges in the left graph, two edges $\text{<}b,\text{L-R}, f\text{>}$ and $\text{<}f,\text{T-B},g\text{>}$ can be deduced and thus should be added to the graph.

It should be clarified the six conditions in Definition~\ref{def:ps} impose no constraints on a human player's exploration activity, since these conditions are the basic essentials for a valid partial solution. However, when the virtual environment tries to modify a player's partial solution (for example, in the feedback activity), it should ensure that the modification will not violate any of the six conditions.

The following notations are used in subsequent sections.

\begin{itemize}
\item $e_L(v,p,t)$: the edge connected to the left side of vertex $v$ in $\mathcal{PS}(p,t)$; that is, $e_L(v,p,t).R = v$. If $v$'s left side connects no edge, then $e_L(v,p,t)$ will return a $null$ value. Similarly, $e_R(v,p,t)$, $e_T(v,p,t)$, and $e_B(v,p,t)$ can be defined.
\item $\mathcal{PS}$: the set consisting of all partial solutions to the puzzle.
\item $C(v,p,t)$: the connected graph in $\mathcal{PS}(p,t)$ that includes vertex $v$. 
\end{itemize}

\begin{definition}[A Player's Rejected Edge Set at Any Time] Given a PJ puzzle, for any player $p \in \mathcal{P}$, at time $t$, the player's rejected edge set, denoted as $\mathcal{R}(p,t)$, is defined as
$$\mathcal{R}(p,t) = \left\{ e \ \bigg| 
\begin{array}{l}
  \forall C \in \mathcal{PS}(p,t) \cdot e \notin \mathcal{E}(C),\\
  \exists u < t \cdot \exists C \in \mathcal{PS}(p,u) \cdot e \in  \mathcal{E}(C)
\end{array} 
\right\}$$
\end{definition}

That is, a player's rejected edge set at time $t$ consists of all those edges that are not included in the player's current partial solution, but were included in the player's partial solution at a previous time.

We use $R(e,p,t)$ to denote the connected graph from which the edge $e$ in $\mathcal{R}(p,t)$ was latest removed by the player $p$ before time $t$.
$$
 \{R(e,p,t)\} = \left\{C \ \bigg| 
\begin{array}{l} 
 e \in \mathcal{E}(C), C \in \mathcal{PS}(p,u), u < t, \\
 
 \forall u < t' \le t \cdot \forall C' \in \mathcal{PS}(p,t') \cdot e \notin \mathcal{E}(C')
\end{array} 
 \right\}
$$

\begin{proposition} Given a PJ puzzle, for any player $p \in \mathcal{P}$ and any time $t$, $ \forall C \in \mathcal{PS}(p,t) \cdot \mathcal{E}(C) \cap \mathcal{R}(p,t) = \varnothing$
\end{proposition}

That is, an edge cannot be included in a player's partial solution and rejected edge set at the same time.

\subsection{Integration}
\label{sec:integration}

At any time, through the integration activity, all players' partial solutions and rejected edge sets are integrated in real time into an artifact called the \emph{collective opinion graph} (COG).

\begin{definition}[Collective Opinion Graph]\label{def:cog} Given a PJ puzzle, the collective opinion graph of the group of human players $\mathcal{P}$ at time $t$, is a graph $Q_t = (\mathcal{V}(Q_t), \mathcal{E}(Q_t))$ that is defined as
\begin{enumerate}
\item $\mathcal{V}(Q_t) = \mathcal{V}$,
\item $\mathcal{E}(Q_t) = \left[\bigcup_{p \in \mathcal{P}, C \in \mathcal{PS}(p,t)} \mathcal{E}(C) \right] \bigcup \left[ \bigcup_{p \in \mathcal{P}}{\mathcal{R}(p,t)} \right]$.
\end{enumerate}
\end{definition}

For every edge $e \in \mathcal{E}(Q_t)$, we maintain two sets of human players: the set of players who support $e$, denoted as $\mathcal{P}_{sup}(e,t)$; and the set of players who reject $e$, denoted as $\mathcal{P}_{rej}(e,t)$.
\begin{itemize}
\item $\mathcal{P}_{sup}(e,t) = \{ p \ | \ p \in \mathcal{P}, \exists C \in \mathcal{PS}(p,t) \cdot e \in \mathcal{E}(C) \}$,
\item $\mathcal{P}_{rej}(e,t) = \{ p \ | \ p \in \mathcal{P}, e \in \mathcal{R}(p,t)\}$.
\end{itemize}

For every edge $e \in \mathcal{E}(Q_t)$, we also maintain two weights: the \emph{positive} weight $w^+(e,t)$, and the \emph{negative} weight $w^-(e,t)$.
\begin{itemize}
\item $w^{+}(e,t) = \sum_{p \in \mathcal{P}_{sup}(e,t), C \in \mathcal{PS}(p,t)}{\mathbf{1}(e \in \mathcal{E}(C)) \cdot |\mathcal{E}(C)|}$,
\item $w^{-}(e,t) = \sum_{p \in \mathcal{P}_{rej}(e,t)}{|\mathcal{E}(R(e,p,t))|}$.
\end{itemize}

Here, $\mathbf{1}(x)$ is an indicator function that returns $1$ if its argument $x$ is true, and $0$ otherwise. It should be noticed that the positive weight of an edge is not the number of players who support the edge, but the sum of the number of edges of all those connected graphs that include the edge in players' current partial solutions; it's a weighted form of the the number of supporting players. It is the similar for the negative weight of an edge.

For an edge $e \in \mathcal{E}(Q_t)$, its \emph{confidence factor} $\varphi(e,t)$ is defined as
$$\varphi(e,t) = \frac{w^+(e,t)}{w^+(e,t) + w^-(e,t)}.$$

For every vertex $v \in \mathcal{V}(Q_t)$, the set of edges connected to the left side of $v$, denoted as $\mathcal{E}_L(v,t)$, is defined as
$$\mathcal{E}_L(v,t) = \{ e \ | \ e \in \mathcal{E}(Q_t), e.R = v \}.$$

The $\phi$-\emph{effective} edge set of $v$ at the left side, denoted as $\mathcal{E}_L(v,t \ | \ \phi)$, is defined as
$$\mathcal{E}_L(v,t \ | \ \phi) = \{ e \ | \ e \in \mathcal{E}_L(v,t), \varphi(e,t) \ge \phi \},$$
where, $\phi$ is a constant ratio value (i.e., a real value in $[0, 1]$).

The $(\phi, \epsilon)$-\emph{strong effective} edge set of $v$ at the left side, denoted as $\mathcal{E}_L(v,t \ | \ \phi, \epsilon)$, is defined as
$$\mathcal{E}_L(v,t \ | \ \phi, \epsilon) = \left\{ e \ \Bigg|  \begin{array}{l} 
e \text{ is the top } k \text{ edges in } \mathcal{E}_L(v,t \ | \ \phi) \\ 
\text{according to their positive weights},\\
k = | \lceil W_L(v,t \ | \ \phi) \rceil^\epsilon |
\end{array} \right\},$$
where, $\epsilon$ is a constant ratio value near or equal to 0, $W_L(v,t \ | \ \phi)$ is the descending-ordered sequence of all the edges in $\mathcal{E}_L(v,t \ | \ \phi)$ according to their positive weights, and $\lceil W_L(v,t \ | \ \phi) \rceil^\epsilon$ is the $\epsilon$-\emph{distinguished prefix} of $W_L(v,t \ | \ \phi)$ (see Definition~\ref{def:epsilon-prefix} in Appendix).

Similarly, we can define the $(\phi, \epsilon)$-\emph{strong effective} edge sets of a vertex $v \in \mathcal{V}(Q_t)$ at the \emph{right}, \emph{top}, and \emph{bottom} sides, denoted as $\mathcal{E}_R(v,t \ | \ \phi)$, $\mathcal{E}_T(v,t \ | \ \phi)$, and $\mathcal{E}_B(v,t \ | \ \phi)$, respectively.

The purpose of $(\phi, \epsilon)$-\emph{strong effective} edge sets is to filter out a set of edges with high probability of being included in the correct solution to the PJ puzzle. These edges will be recommended to players at appropriate time through the feedback activity.

\subsection{Feedback}
\label{sec:feedback}
Through the feedback activity, player-specific recommendations are calculated for each player, according to the current COG and a player's current partial solution, in order to improve players' solving efficiency. We focus on three aspects of feedback: \emph{when}, \emph{what}, and \emph{how}.   

\subsubsection{When, What, and How} Two kinds of time point (\emph{when}) of feedback are identified: after-operation, and into-stagnation. The \emph{after-operation} denotes those time points that are just after the finish of some operations conducted by players (for example, connecting two pieces together). The \emph{into-stagnation} denotes those time points players enter into the state of stagnation (that is, the player' partial solution and rejected edge set stays unchanged for at least a predefined time period).

The contents (\emph{what}) of feedback to a player are one or more edges that are not included in the player's current partial solution, but in certain $(\phi, \epsilon)$-\emph{strong effective} edge sets. 

We employ two ways (\emph{how}) to recommend the contents of feedback to a player: connecting-action and edge-hint. Recommending an edge through \emph{connecting-action} means that the two pieces in the edge are automatically connected together in the player's current partial solution. Recommending an edge through \emph{edge-hint} means that the two corresponding sides of two pieces in the edge are highlighted with the same distinct color in the player's workspace.  

Based on different combinations of \emph{when}, \emph{what}, and \emph{how}, we design two kinds of feedback policy: \emph{responsive}, and \emph{stimulative}.

\subsubsection{Responsive Feedback}
Responsive feedback is triggered at the after-operation time points. Given such a time point of a player, the \emph{what} and \emph{how} of feedback are decided as follows.

First, obtain the focused connected graph of the operation, that is, the connected graph the player's mouse cursor is on when the operation is finished. Then, for each of the vertices in the focused connected graph whose \emph{degree} is less than 4, and for each of the empty sides (i.e., sides connecting no edge) of the vertex, check its $(\phi, \epsilon)$-\emph{strong effective} edge set: if this set contains only one edge, and adding the edge into the player's current partial solution will not cause violation to Definition~\ref{def:ps}, then this edge is recommended through connecting-action (as a result, the player's current partial solution is changed). If after traversing, no edge is recommended, then randomly select a $(\phi, \epsilon)$-\emph{strong effective} edge set being traversed, and all the edges in the set are recommended through edge-hint (this will not change the player's current partial solution).

\subsubsection{Stimulative Feedback} Stimulative feedback is triggered at the into-stagnation time points. Give such a time point $t$ of a player, the \emph{what} and \emph{how} of feedback are decided as follows.

First, get all \emph{best-buddy} edges related to the player's current partial solution. A best-buddy edge is an edge, for example, $\text{<}u, \text{L-R}, v\text{>}$, that satisfies two conditions: (1) this edge is not included in the player's partial solution; (2) $\mathcal{E}_R(u,t \ | \ \phi, \epsilon) = \mathcal{E}_L(v,t \ | \ \phi, \epsilon) = \{\text{<}u, \text{L-R}, v\text{>}\}$. Then, traverse all the best-buddy edges using a random order: for each best-buddy edge, if adding it into the player's current partial solution will not cause violation to Definition~\ref{def:ps}, then this edge is recommended through connecting-action. If after traversing, no edge is recommended, then randomly select a best-buddy edge and recommend it through edge-hint.

\section{Experiment and Evaluation}

In this section, we evaluate the proposed approach through a set of controlled experiments. Three research questions are particularly focused, and statistical findings about the three questions are presented and briefly analyzed.

\subsection{Research Questions}
\label{sec:rqs}

\begin{itemize}
\item \textbf{RQ1: Collective Performance}

\textit{How do the two factors of puzzle size and group size influence the collective performance of PJ puzzle solving?}

\item \textbf{RQ2: Feedback Precision and Ratio}

\textit{How do the two factors of puzzle size and group size influence feedback precision and feedback ratio in PJ puzzle solving?}

\item \textbf{RQ3: Stigmergy-based Collaboration vs. Face-to-Face Collaboration and Automatic PJ Puzzle Solvers}

\textit{Will stigmergy-based collaboration outperform face-to-face collaboration and automatic solvers in PJ puzzle solving?}

\end{itemize}

\subsection{Experiment Platform}

\begin{figure}[!htp]
\includegraphics[width=1\linewidth]{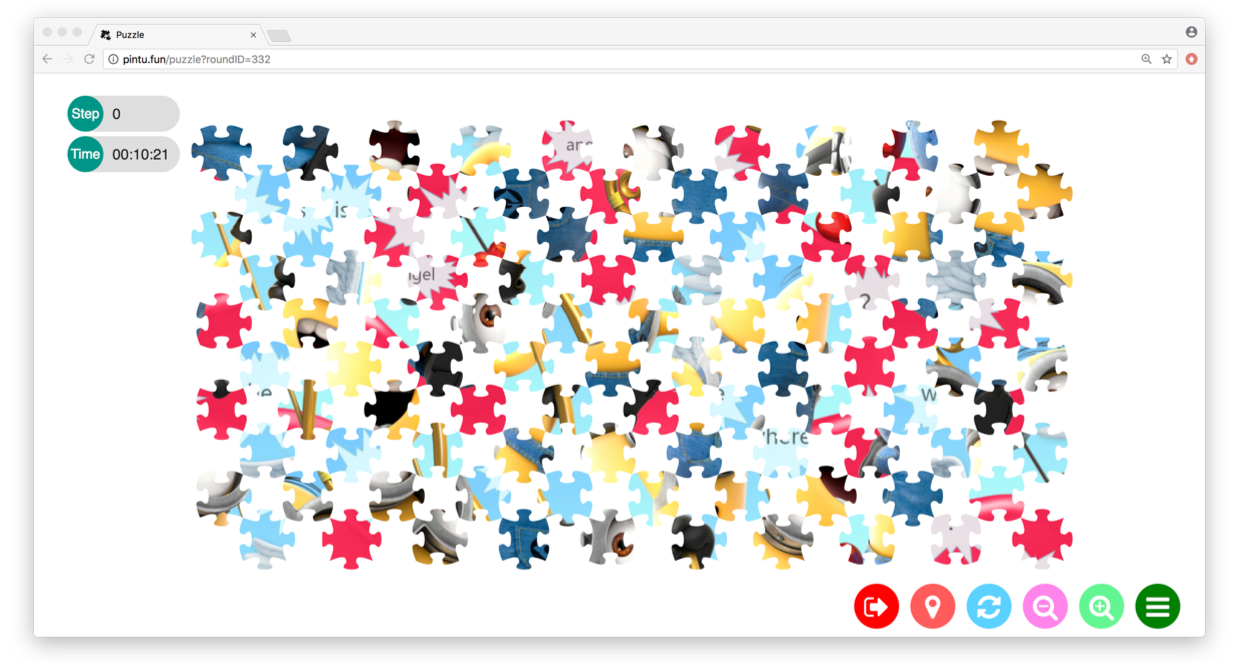}
\caption{Screenshot of a player's puzzle-solving workspace.}
\label{fig:cjp}
\end{figure}

The experiments are carried out on \emph{Crowd Jigsaw Puzzle}, an on-line platform (available at http://www.pintu.fun) developed to support the proposed approach. The platform runs on a web server with 4-core CPU, 8GB RAM, and CentOS 7. The experiments are organized as a set of game rounds, each of which consists of a PJ puzzle and a player group. In a game round, each player tries to solve the puzzle in her/his own workspace. Fig.~\ref{fig:cjp} shows the screenshot of a player's workspace at the beginning of a game round.

\subsection{Method}

\subsubsection{Participants}

Fifty-two participants (aged range: 19-45 years; $M_{age} = 24.23$; $SD_{age} = 21.87$; 30 males) were recruited through campus BBS and social networks. Among them, 32 (61.5\%) are postgraduates, 15 (28.9\%) are undergraduates, and 5 (9.6\%) are college staff. Payments were contingent on participants' performance, consisting of a base payment and a bonus payment. At the beginning of a game round, each participant received a fixed amount of money as the base payment, and after the game round, each participant further received a varying amount of bonus according to her/his performance in the game round.

\subsubsection{Experimental Procedure}
\label{sec:tasks}

Before the experiment, each participant was asked to register an account, take a tutorial and finish at least one PJ puzzle on the platform to get familiar with the game environment. In the experiment, each participant is required to perform following tasks in sequence: (1) sign in, waiting for a new game round; (2) when the number of players reaches the pre-assigned group size of a game round, begin to solve the puzzle in her/his own workspace; (3) when any player resolves the puzzle, complete a questionnaire about the puzzle-solving process, and then quit the current game round.

\subsubsection{Parameter Settings} All pieces in a PJ puzzle game have the same jagged-square shape, including pieces at the borders of the picture to be recovered. The outermost 2 row/column pixels of each piece are erased. The puzzle size (\emph{ps}) ranges from 4$\times$4 to 10$\times$10, and the group size (\emph{gs}) from 1 to 10; as a result, there are totally 70 different combinations of \emph{ps} and \emph{gs}. More than 80 images are collected as the to-be-recovered pictures. The two parameters of $\phi$ and $\epsilon$ (see Section~\ref{sec:integration}) are assigned with $0.618$ and $0.02$, respectively.

\subsection{Task and Design}

\subsubsection{RQ1: Collective Performance.}
\label{exp:rq1}

The \emph{ps} and \emph{gs} are taken as independent variables, and the collective performance \emph{cp} as the induced variable. The \emph{cp} of a game round is defined as the time of the best performing player to solve the PJ puzzle (measured in seconds) in the player group. For each combination of \emph{ps} and \emph{gs}, 5 different game rounds were carried out. Totally 350 game rounds were carried out in 3 months by the 52 participants.

\begin{figure}[!htp]
\includegraphics[width=0.8\linewidth]{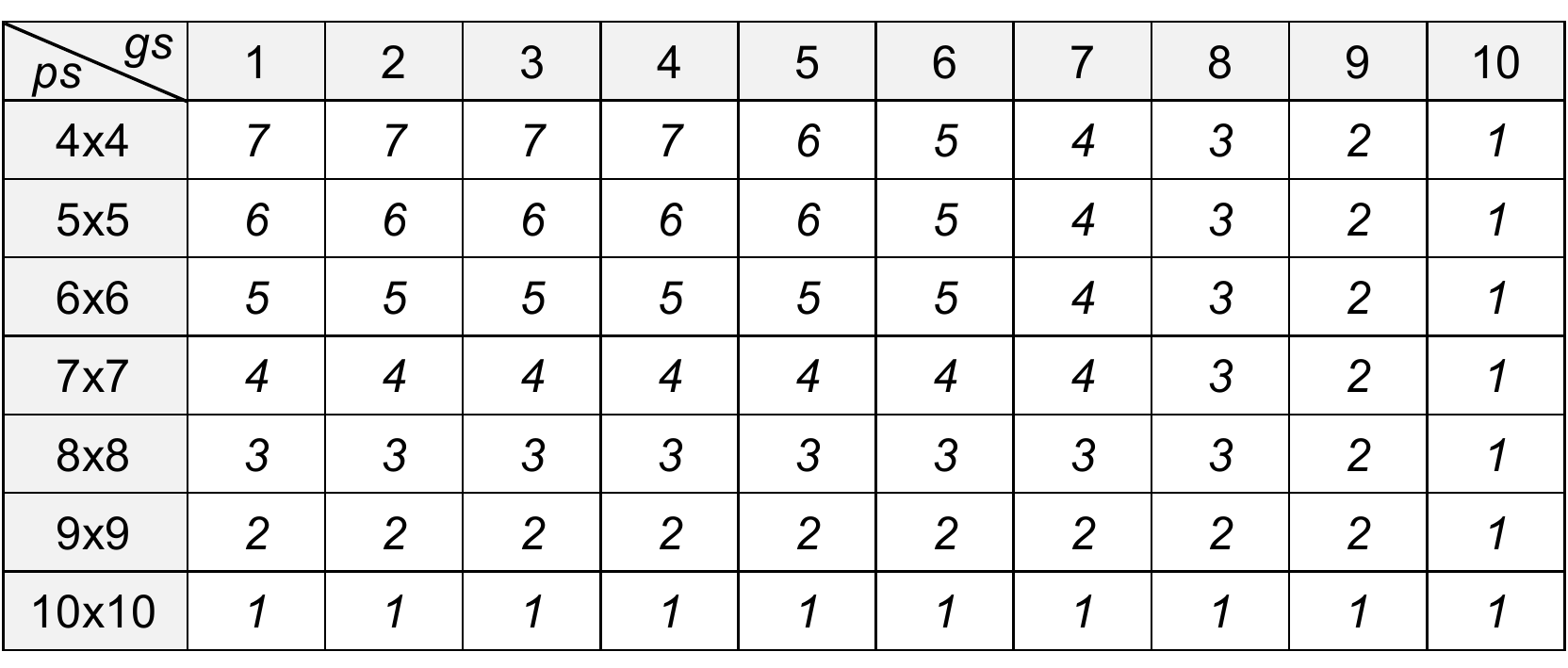}
\caption{The 7 batches of 350 game rounds. The number in a cell (i.e., a combination of \emph{ps} and \emph{gs}) is the batch number.}
\label{fig:7-batch}
\end{figure}

The 350 game rounds were partitioned into 7 batches (see Fig.~\ref{fig:7-batch}). Game rounds in \emph{batch 1} were carried out firstly, then that in \emph{batch 2}, \emph{3}, until \emph{batch 7}. For all the game rounds in \emph{batch} $i \in [1, 7]$, $(18-2 \cdot i) \cdot 5$ different pictures were randomly selected, each of which served as the picture in a game round. For every 5 game rounds with same \emph{ps} and \emph{gs}, 5 groups of player were randomly selected without replacement, each of which served as the player group in a game round. 

The 7-batch based process is designed to alleviate the influence of players' familiarity with pictures on \emph{cp}. If a player firstly participates in a game round as a single-player group, and then in a game round with the same picture but in a 10-player group, the player's familiarity with the picture will make the \emph{cp} observed in the latter round higher than its real value. The 7-batch based process ensures that if a picture appears in two game rounds, the latter will always have a \emph{gs} value less than the former. Therefore, the experiment results, although not accurate absolutely, are an \emph{underestimation} of the improvement of \emph{cp} brought by increasing the \emph{gs} value.

\subsubsection{RQ2: Feedback Precision and Ratio.}

\begin{itemize}
\item \emph{Feedback Ratio}. Given a game round, the feedback ratio \emph{fr} evaluates the percentage of edges recommended through connecting-action in the best performing player's solution.
\item \emph{Feedback Precision}. The feedback precision \emph{fp} evaluates the percentage of correctly recommended edges in all the recommended edges through connecting-action in the best performing player's solving process.
\end{itemize}

Besides the two quantitative metrics, we also analyzed players' qualitative evaluations to the feedback activities in their solving processes. These qualitative evaluations were collected through a questionnaire before a player quited a game round. 

\subsubsection{RQ3: Stigmergy-based Collaboration vs. Face-to-Face Collaboration and Automatic PJ Puzzle Solvers.}
To compare our approach with traditional collaborative ways to solve PJ puzzle, we conducted a set of experiments, in which player groups tried to solve PJ puzzle through face-to-face collaboration. Five different 10-player groups were randomly selected. For each of them, 7 game rounds were carried out, each of which involved a different picture and a distinct \emph{ps} $\in [$4$\times$4$, $ 10$\times$10$]$. In each game round, 10 players sat before a projector screen, on which a PJ puzzle-solving workspace was displayed; one player was responsible for manipulating pieces in the workspace, and others kept telling their opinions to the manipulator.

To compare our approach with automatic solvers, we selected the state-of-the-art solver for PJ puzzle with known piece-orientation \cite{sholomon2013genetic, sholomon2016automatic}, and conducted a set of experiments that used the automatic solver to solve the same set of 10-player PJ puzzles in Section~\ref{exp:rq1}. The automatic solver was run on a desktop computer with 3.6GHz Intel Core i7 CPU, 16GB RAM, and Ubuntu 18.04.

\subsection{Results and Analysis}
\label{sec:results}

\subsubsection{RQ1: Collective Performance}

\begin{figure}
\includegraphics[width=1\linewidth]{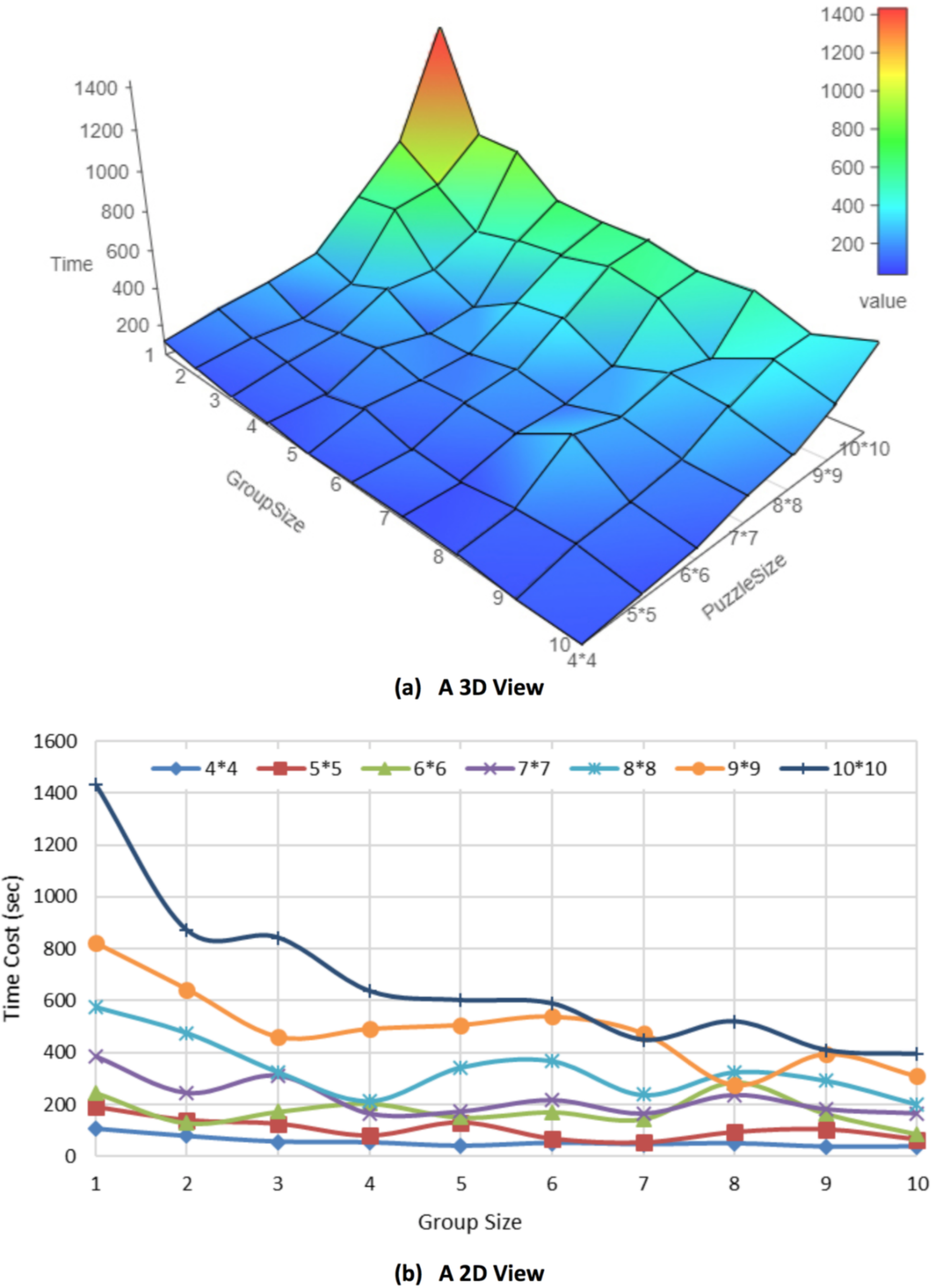}
\caption{The average time for player groups with \emph{gs} $\in [1, 10]$ to solve PJ puzzle with \emph{ps} $\in [$4$\times$4$, $10$\times$10$]$.}
\label{fig:rq1-time}
\end{figure}

Fig.~\ref{fig:rq1-time} shows the time for a player group with \emph{gs} $\in [1, 10]$ to solve PJ puzzles with \emph{ps} $\in [$4$\times$4$, $10$\times$10$]$. The time of each combination of \emph{ps} and \emph{gs} is an average value of the 5 game rounds for this combination. From the results, it can be observed \emph{in general} that: (1) given a PJ puzzle with a fixed \emph{ps} value, the puzzle-solving time decreases with the increasing of the \emph{gs} value, and (2) given a PJ puzzle with a fixed \emph{gs} value, the puzzle-solving time increases with the increasing of the \emph{ps} value.

Table~\ref{tab:improvement} shows a quantitative analysis to the time improvement brought by playing as a group, comparing with the best single players. Column 1 gives the best single-player's time to solve PJ puzzle with \emph{ps} $\in [$4$\times$4$, $10$\times$10$]$. For example, the value of 385.38 in row 4 of column 1 means that, in the 5 game rounds with \emph{ps}=7$\times$7 and \emph{gs}=1, the minimum puzzle-solving time is 385.38 seconds. Column $i \in [2, 10]$ shows the average time improvement when a PJ puzzle game is played by a $i$-player group, comparing with the corresponding time value in column 1. The last row in column $i$ shows the average time improvement at all 7 \emph{ps} values. It can be observed that, \emph{on average, the time improvement brought by playing as a group ranges from 31.36\% to 64.57\%, as the \emph{gs} value ranges from 2 to 10}.

\newcommand{\tabincell}[2]{\begin{tabular}{@{}#1@{}}#2\end{tabular}}  
\begin{table*}
\begin{threeparttable}
  \caption{ Time improvement brought by playing as a group, comparing with the best single players}
  \label{tab:improvement}
  \begin{tabular}{cccccccccccc}
    \toprule
    \backslashbox{\textit{ps}}{\textit{gs}} & \tabincell{c}{1\\(second)} & 2 & 3 & 4 & 5 & 6 & 7 & 8 & 9 & 10 \\
    \midrule
    4*4 & 108.12 & 26.93\% &47.51\% &50.05\% &63.01\% &51.91\% &56.99\% &53.76\% &65.78\% &64.86\% \\
    5*5 & 191.50 & 26.89\% &34.20\% &58.23\% &31.59\% &64.49\% &72.32\% &51.18\% &45.43\% &65.83\% \\
    6*6 & 244.67 & 47.28\% & 29.56\% &16.83\% &38.49\% &30.11\% &41.14\% &16.49\% &32.56\% &64.50\% \\
    7*7 & 385.38 & 36.69\% & 19.50\% &57.44\% &55.37\% &43.95\% &57.44\% &38.76\% &53.03\% &57.23\%  \\
    8*8 & 575.18 & 20.92\% &43.32\% & 62.62\% &40.31\% &36.02\% &58.62\% &43.50\% &49.06\% &64.88\% \\
    9*9 & 821.17 & 21.58\%& 43.80\% &40.07\% &38.25\% &34.35\% &42.14\% &66.26\% &51.89\% &62.16\% \\
    10*10 & 1432.12 & 39.20\% &41.17\% &55.56\% &58.03\% &58.90\% &68.68\% & 63.72\% &71.44\% &72.51\%\\
    Average & 536.88 & 31.36\% & 37.01\% & 48.69\% & 46.44\% & 45.68\% & 56.76\% & 47.67\% & 52.74\% & 64.57\% \\
  \bottomrule
\end{tabular}
\end{threeparttable}
\end{table*}

\begin{figure}[!htp]
\includegraphics[width=1\linewidth]{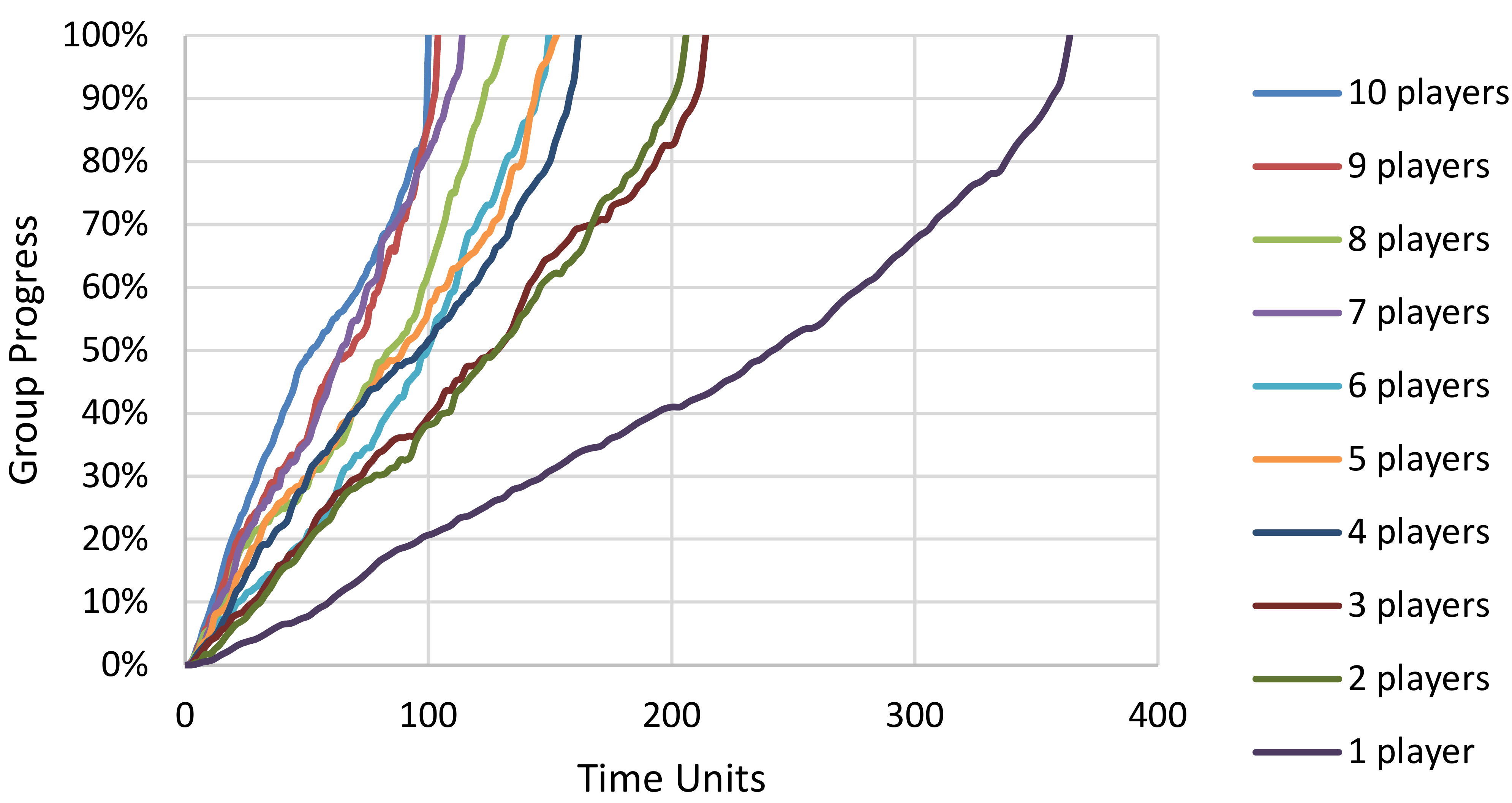}
\caption{The puzzle-solving progress of a player group with \emph{gs} $\in [1, 10]$ to solve a 10$\times$10 PJ puzzle. One time unit is defined as 1\% of the average puzzle-solving time of 10-player groups.}
\label{fig:rq1-progress}
\end{figure}

Fig.~\ref{fig:rq1-progress} shows the puzzle-solving progress for player groups with \emph{gs} $\in [1, 10]$ to solve a 10$\times$10 PJ puzzle. In this figure, the average puzzle-solving time of 10-player groups is normalized as 100. At any time, a group's puzzle-solving progress is defined as the ratio of \emph{the number of correct edges appearing in any $(\phi, \epsilon)$-strong effective edge set of the current COG} to \emph{the number of edges in the correct solution}, a value between 0 and 1. Two observations could be found from the 10 progress curves: (1) the puzzle-solving progress looks like a linear function of the time, except when its value  is greater than 0.85; (2) the slope coefficient has a significant increase when $gs$ changes from 1 to 2 or 3, which means \emph{even playing as a small group with only 2 or 3 players, the puzzle-solving process could be accelerated significantly (nearly 50\% acceleration in our experiments)}. 

In order to obtain an accurate quantitative relation between the puzzle-solving time \emph{cp} (namely, the collective performance) and the two factors of \emph{gs} and \emph{ps}, we identify 3 kinds of function $cp = f(ps, gs)$, based on our intuitive understanding of this function.
$$
\begin{array}{l}
\text{(1)} \ \ cp = f(ps, gs \ | \ \mu, \upsilon) = \mu \cdot e^{\upsilon \cdot ps} \cdot gs^{-1} \vspace{1ex}\\
\text{(2)} \ \ cp = f(ps, gs \ | \ \mu, \upsilon, \omega) = \mu \cdot e^{\upsilon \cdot ps} \cdot (gs + \omega)^{-1} \vspace{1ex}\\
\text{(3)} \ \ cp = f(ps, gs \ | \ \mu, \upsilon, \omega) = \mu \cdot e^{\upsilon \cdot ps} \cdot e^{-\omega \cdot gs} \\
\end{array}
$$
The term $\mu \cdot e^{\upsilon \cdot ps}$ is based on the fact that the jigsaw puzzle \emph{in general} is a NP-complete problem, thus \emph{cp} will increase exponentially as \emph{ps} increases. The term $gs^{-1}$ shows a perfect linear collaboration among players: if a single-player group's \emph{cp} is $t$, then a $n$-player group's \emph{cp} will be $t/n$. The term $(gs+\omega)^{-1}$ shows a variant of the perfect linear collaboration by adding a constant $\omega$: Whether $\omega$ is positive (or negative) indicates whether the collaboration is sub-linear (or super-linear). The term $e^{-\omega \cdot gs}$ shows that \emph{cp} will decrease exponentially as \emph{ps} increases, indicating the fastest kind of super-linear collaboration. 

For each of the 3 kinds of function, we use linear regression to get the concrete function having a maximum fitness to the experiment results, and the corresponding $r^2$ (\emph{coefficient of determination}) value. 
$$
\begin{array}{ll}
\text{(1)} \ \ 39.661 \cdot e^{0.381 \cdot ps}  \cdot  gs^{-1} , & r^2 = 0.6417  \vspace{1ex}\\
\text{(2)} \ \ 149.50 \cdot e^{0.362 \cdot ps}  \cdot (gs + 3.391)^{-1}, & r^2 = 0.8982 \vspace{1ex}\\
\text{(3)} \ \ 36.307 \cdot e^{0.361 \cdot ps} \cdot e^{-0.130 \cdot gs}, & r^2 = 0.8893 \\
\end{array}
$$
Based on the former two functions, we believe that \emph{in our experiments we observe the sub-linear collaboration among groups with 2 to 10 players}. But considering the last function, \emph{we don't know whether the collaboration will keep sub-linear or become super-linear for groups with more than 10 players}. 




\subsubsection{RQ2: Feedback Precision and Ratio}

\begin{figure}[!htp]
\includegraphics[width=1\linewidth]{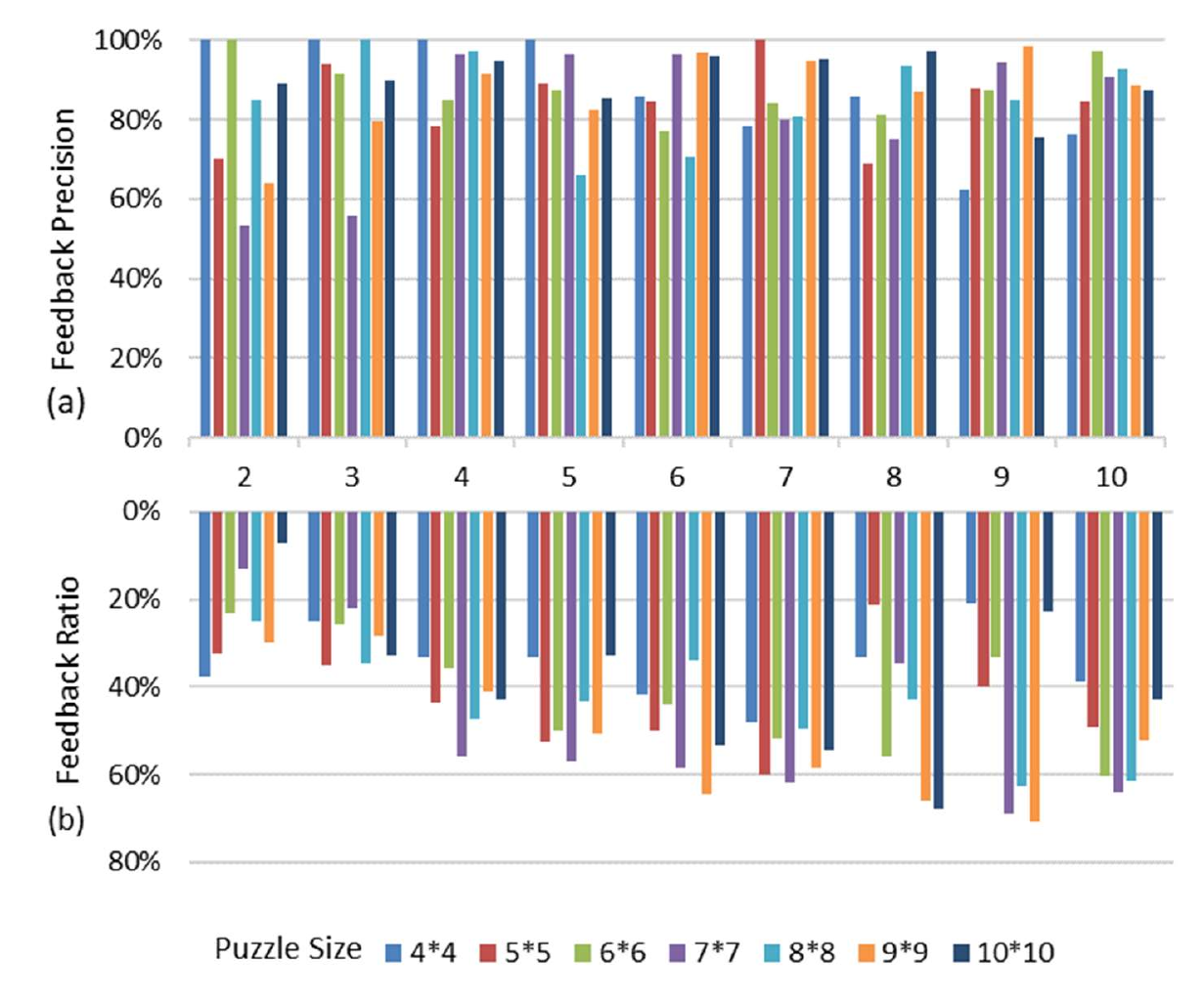}
\caption{Average feedback precision and feedback ratio for different combinations of puzzle size and group size.}
\label{fig:ratio_pre}
\end{figure}

Fig.~\ref{fig:ratio_pre} shows the average \emph{fp} and \emph{fr} values for different combinations of \emph{ps} and \emph{gs}. \emph{Our approach obtains a mean fp of 86.34\% with the standard deviation of 11.22\%, which indicates that the integration and feedback mechanism in our approach is relatively effective and stable}. 

\emph{The fr value shows a trend of increasing in general as gs increases}. For \emph{gs} $\in [8, 10]$, the mean \emph{fr} is around 45\%, which means that in the correct solution of the best performing player, about 45\% edges come from other players' partial solutions. However, we do observe that for \emph{gs} $\in [8, 10]$ the standard deviation of \emph{fr} becomes bigger than that of \emph{gs} $\in [4, 7]$, and currently we do not have convincing explanations to this observation.  

\begin{figure}[!htp]
\includegraphics[width=0.7\linewidth]{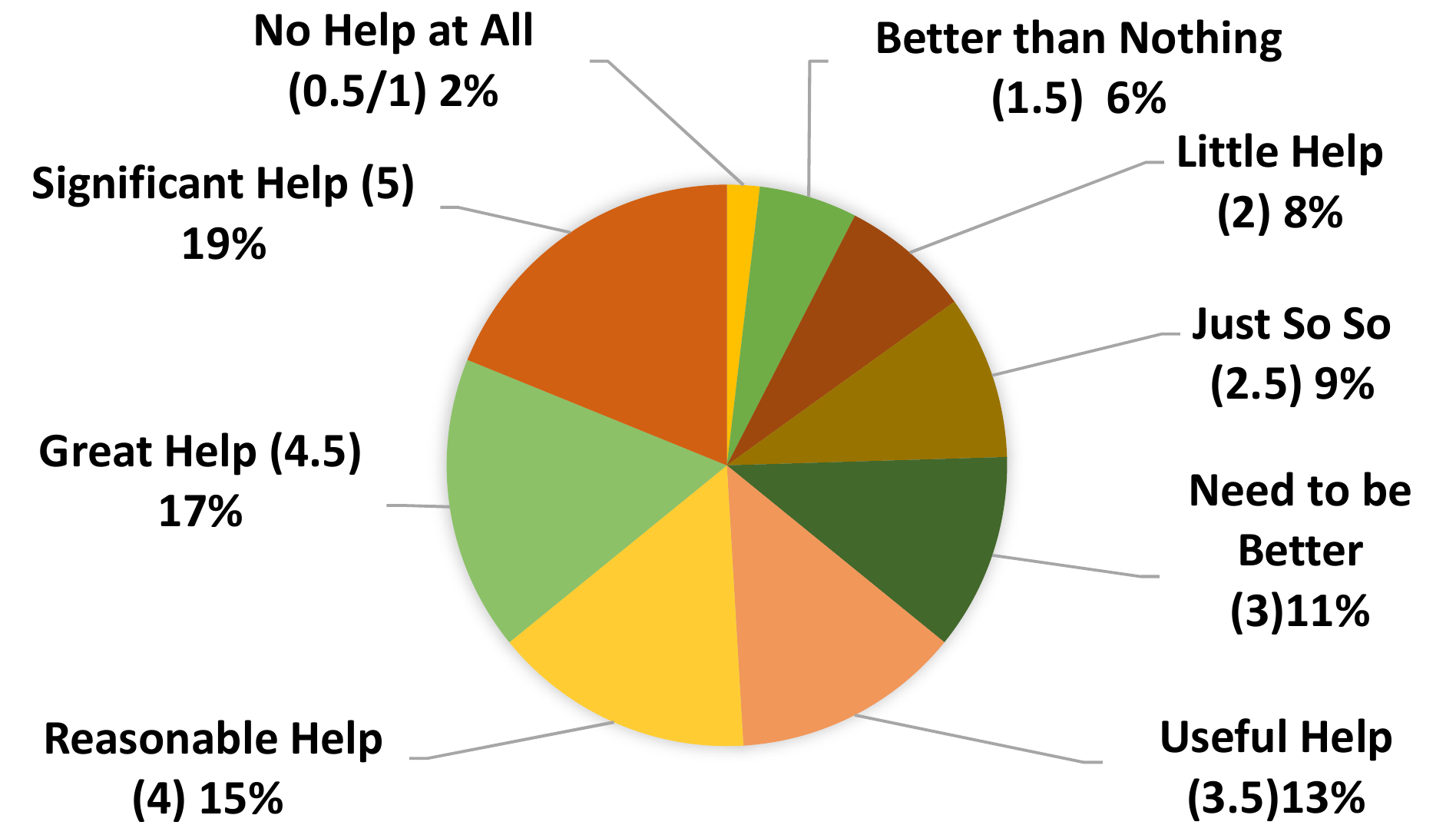}
\caption{Qualitative evaluations from players about the feedback information received in PJ puzzle solving. }
\label{fig:ratings}
\end{figure}

Fig.~\ref{fig:ratings} shows the qualitative evaluations from players about the feedback information they have received in the puzzle-solving process. \emph{64\% of players give a rating value over 3.5 (\emph{useful help})}, which is consistent with the quantitative evaluation. 

\subsubsection{RQ3:  Stigmergy-based Collaboration vs. Face-to-Face Collaboration and Automatic PJ Puzzle Solvers.}

\begin{figure}[!htp]
 \includegraphics[width=1\linewidth]{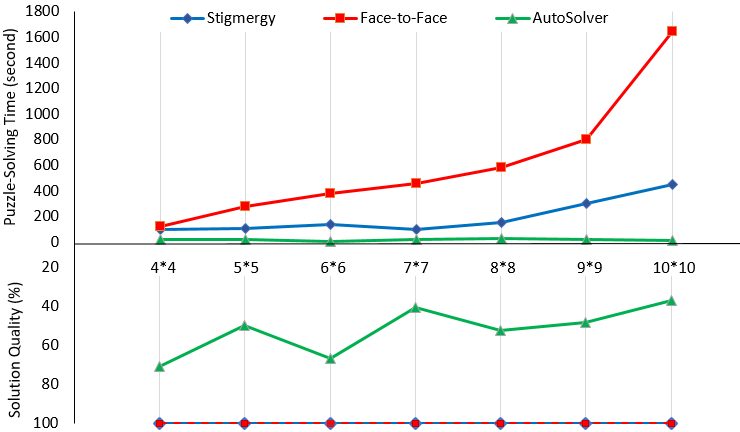}
\caption{Puzzle-solving time and solution quality of stigmergy-based collaboration (our approach), face-to-face collaboration, and an automatic solver. }
\label{fig:stig-vs-face_solver}
\end{figure}

Fig.~\ref{fig:stig-vs-face_solver} shows the comparison of puzzle-solving time and solution quality among stigmergy-based collaboration (our approach), face-to-face collaboration, and the automatic solver. The solution quality is defined as the percentage of correct edges in a candidate solution to PJ puzzle. Among the three approaches, the automatic solver shows minimum puzzle-solving time; for 10$\times$10 PJ puzzles, the solver can find a candidate solution in only 11 seconds. However, \emph{the automatic solver shows a relatively lower solution quality: it has a mean solution quality of 0.52, while that value of the other two approaches both are 1}. In both of the two collaboration-based approaches, the solving time increases with the increasing of puzzle size. However, the face-to-face collaboration has a more rapid increasing than stigmergy-based collaboration. That is, \emph{for 10-player groups, our approach shows a better scalability to puzzle size than face-to-face collaboration}.

\section{Discussion and Future Work}

In this section, we briefly discuss some elementary problems related to the proposed approach, and highlight some of our future work. 

\emph{Threats to validity.} Our experiments involve only a small number (52) of human subjects who are mainly college students. Although we have adopted several methods (like repeating each combination of \emph{puzzle size} and \emph{group size} for 5 times, and alleviating the influence of players' familiarity with pictures on \emph{collective performance} by using a 7-batch based process) to minimize the possible biases and deviations, it nearly impossible for us to ensure that there is no sampling bias and statistical deviation in the experiment results. 

\emph{Scalability to problem size and group size.} One essential characteristic of stigmergy-based CI is its good scalability to \emph{group size}. Our approach's scalability to \emph{group size} depends on an important factor, that is, the computing capability of the server that supports the virtual environment with integration and feedback mechanisms. As long as the server could process and response a player's operation without obvious delay, the approach would work well. Our approach's scalability to \emph{puzzle size} depends on the size and resolution of the computer display used by the player.

\emph{Quantitative evaluation of CI.} We think that human CI systems could be evaluated by their scalability to group size and problem size. That is, given two human CI systems addressing the same kind of problems, it can be quantitatively evaluated that which system is better or more intelligent than the other, according to the decreasing speed of problem-solving time as the group size increases, and the increasing speed of problem-solving time as the puzzle size increases.

\emph{The applicability of the EIF loop to other problems.} The EIF loop clarifies two implicit responsibilities of the environment in stigmergy, and points out an engineering framework for artificial CI systems in the cyberspace. We think that, whether the EIF loop could be used to solve a general set of complex problems depends on whether the information pieces possessed by individuals about a problem could be efficiently integrated and then recommended to related individuals; in particular, whether a general mechanism for information integration and feedback could be identified.

Based on the above discussion, our future work will be conducted in two different senses. In the \emph{narrow} sense, we will continue our research on CI-based PJ puzzle solving. We plan to recruit more human players to participate in our experiments to resolve more complex PJ puzzles, in order to empirically examine the scalability of our approach. In the \emph{broad} sense, we will extend our approach to cope with more complex problems in practical situations. Currently, we have located two kinds of complex problem: the \emph{knowledge-graph construction} problem, and the \emph{software development} problem. In addition, we plan to develop a general integration and feedback platform based on graph-based representation of information, as suggested by Romero and Valdez \cite{Romero14}. 

\section{Conclusion}

In this paper, we present an approach to solving PJ puzzle by stigmergy-inspired Internet-based human collective intelligence, that is, by a set of physically distributed human players through a collaborative and decentralized way. The core of this approach is a continuously executing loop, named the \emph{EIF} loop, which consists of three asynchronously connected activities: exploration, integration, and feedback. The key artifact generated by the EIF loop is a continuously-updated \emph{collective opinion graph} (COG), which integrates every human players' opinions in a structured way and in real time, and also serves as the input to the feedback activity. We have developed an on-line platform to demonstrate this approach, and conducted a set of controlled experiments on the platform to investigate the feasibility and effectiveness of this approach. Experiments show that: (1) supported by this approach, the time to solve PJ puzzle is nearly linear to the reciprocal of the number of players, and shows better scalability to puzzle size than that of face-to-face collaboration for 10-player groups; (2) for groups with 2 to 10 players, the puzzle-solving time decreases 31.36\%-64.57\% on average, compared with the best single players in the experiments.

\begin{acks}

The authors would like to thank Dr. Maura Turolla of Telecom
Italia for providing specifications about the application scenario.

The work is supported by the \grantsponsor{GS501100001809}{National
  Natural Science Foundation of
  China}{http://dx.doi.org/10.13039/501100001809} under Grant
No.:~\grantnum{GS501100001809}{61273304\_a}
and~\grantnum[http://www.nnsf.cn/youngscientists]{GS501100001809}{Young
  Scientists' Support Program}.

\end{acks}

\bibliographystyle{ACM-Reference-Format}
\bibliography{sample-bibliography}

\input{sample-bibliography.bbl}
\appendix
\section{Appendix}

\subsection{Notations}
\begin{itemize}
\item $(a_i)_1^K$: the number sequence $a_1, a_2, ..., a_{K}$.
\item $|\mathcal{S}|$: the number of elements in a set or number sequence  $\mathcal{S}$.
\item $\mathcal{V}(G)$: the vertex set of graph $G$.
\item $\mathcal{E}(G)$: the edge set of graph $G$.
\item $d(v,G)$: the degree of vertex $v$ in graph $G$.
\item $\mathcal{V}(e)$: the set of two vertices involved in edge $e$ in a graph.
\item $\mathbf{1}(x)$: an indicator function that is $1$ if its argument $x$ is true, and $0$ otherwise.
\end{itemize}

\subsection{Definitions}

\begin{definition}[The $\epsilon$-Distinguished Prefix of a Decreasing-Ordered Finite Number Sequence] \label{def:epsilon-prefix} Given a decreasing-ordered finite number sequence $(a_i)_1^K$ and a constant $\epsilon$, the $\epsilon$-\emph{distinguished prefix} of this number sequence, denoted as $\lceil (a_i)_{1}^{K} \rceil^\epsilon$, is defined as
\begin{enumerate}
\item $\forall i \in [1, K) \cdot (a_i - a_{i+1}) \le \epsilon \times |a_i| \Rightarrow \lceil (a_i)_{1}^{K} \rceil^\epsilon= (a_i)_{1}^{K}$,
\item $\exists i \in [1, K) \cdot (a_i - a_{i+1}) > \epsilon \times |a_i| \Rightarrow \lceil (a_i)_{1}^{K} \rceil^\epsilon = (a_i)_{1}^{J} \land J \in [1, K) \land (\forall k \in [1, J) \cdot (a_k - a_{k+1}) < (a_J - a_{J+1})) \land (\forall k \in [J+1, K) \cdot (a_k - a_{k+1}) \le (a_J - a_{J+1}))$.
\end{enumerate}
\end{definition}

According this concept, a decreasing-ordered finite number sequence is partitioned into two sequences at the point of two neighbor numbers (for example, $a_i$ and $a_{i+1}$) that has the firstly-appeared maximum difference $(a_i - a_{i+1})$ in the sequence, unless for each two neighbors $a_j$ and $a_{j+1}$, the relative difference $\frac{a_j - a_{j+1}}{a_j}$ is less than or equals to $\epsilon$. In the latter case, the $\epsilon$-\emph{distinguished prefix} of a decreasing-ordered finite number sequence is the sequence itself.

For example, given a decreasing-ordered finite number sequence <$10,9,8,7,6,3,2,1$>, for any $\epsilon < 0.5$, the $\epsilon$-\emph{distinguished prefix} of this number sequence is the sequence of <$10,9,8,7,6$>; and for any $\epsilon \ge 0.5$, the $\epsilon$-\emph{distinguished prefix} of this number sequence is the sequence itself. Given another sequence <$10,9.9,9.8,9.7$>, for any $\epsilon < \frac{1}{98}$, the $\epsilon$-\emph{distinguished prefix} is the <$10$>; and for any $\epsilon \ge \frac{1}{98}$, the $\epsilon$-\emph{distinguished prefix} is the sequence itself.